\documentclass{article} % For LaTeX2e
\usepackage{iclr2021_conference,times}

\usepackage{amsmath,amsfonts,bm}

\def\eqref#1{equation~\ref{#1}}
\def\1{\bm{1}}

\def\eps{{\epsilon}}

\DeclareMathAlphabet{\mathsfit}{\encodingdefault}{\sfdefault}{m}{sl}
\SetMathAlphabet{\mathsfit}{bold}{\encodingdefault}{\sfdefault}{bx}{n}

\usepackage{hyperref}
\usepackage{url}
\usepackage{float}
\usepackage{wrapfig}
\usepackage{xcolor}

\newcommand{\btheta}{\mathbf{\theta}}

\usepackage{amssymb}

\usepackage{graphicx}
\usepackage[english]{babel}
\usepackage{comment}
\usepackage[T1]{fontenc}

\def\<{\langle}
\def\>{\rangle}

\def\shownotes{1}  %set 1 to show author notes
\ifnum\shownotes=1
\newcommand{\authnote}[2]{$\ll$\textsf{\footnotesize #1 notes: #2}$\gg$}
\else
\newcommand{\authnote}[2]{}
\fi

\title{Influence Functions in Deep Learning \\Are Fragile}

\author{Samyadeep Basu\thanks{Authors contributed equally}, Phillip Pope \footnotemark[1]  \& Soheil Feizi \\
Department of Computer Science\\
University of Maryland, College Park\\
\texttt{\{sbasu12,pepope,sfeizi\}@cs.umd.edu} 
}

\iclrfinalcopy % Uncomment for camera-ready version, but NOT for submission.
\begin{document}

\maketitle

\begin{abstract}
Influence functions approximate the effect of training samples in test-time predictions and have a wide variety of applications in machine learning interpretability and uncertainty estimation. A commonly-used (first-order) influence function can be implemented efficiently as a post-hoc method requiring access only to the gradients and Hessian of the model. For linear models, influence functions are well-defined due to the convexity of the underlying loss function and are generally accurate even across difficult settings where model changes are fairly large such as estimating group influences.
Influence functions, however, are not well-understood in the context of deep learning with non-convex loss functions. 
In this paper, we provide a comprehensive and large-scale empirical study of successes and failures of influence functions in neural network models trained on datasets such as Iris, MNIST, CIFAR-10 and ImageNet.
Through our extensive experiments, we show that the network architecture, its depth and width, as well as the extent of model parameterization and regularization techniques have strong effects in the accuracy of influence functions. In particular, we find that (i) influence estimates are fairly accurate for shallow networks, while for deeper networks the estimates are often erroneous; (ii) for certain network architectures and datasets, training with weight-decay regularization is important to get high-quality influence estimates; and (iii) the accuracy of influence estimates can vary significantly depending on the examined test points. These results suggest that in general influence functions in deep learning are fragile and call for developing improved influence estimation methods to mitigate these issues in non-convex setups.
\end{abstract}

\section{Introduction}
\vspace{-0.6em}
In machine learning, influence functions \citep{cook_influence} can be used to estimate the change in model parameters when the empirical weight distribution of the training samples is perturbed infinitesimally. This approximation is cheaper to compute compared to the expensive process of repeatedly re-training the model to retrieve the exact parameter changes. Influence functions could thus be used to understand the effect of removing an individual training point (or, groups of training samples) on the model predictions at the test-time. Leveraging a first-order Taylor's approximation of the loss function, \citep{influence1} has shown that a (first-order) influence function, computed using the gradient and the Hessian of the loss function, can be useful to interpret machine learning models, fix mislabelled training samples and create data poisoning attacks.

Influence functions are in general well-defined and studied for models such as logistic regression \citep{influence1}, where the underlying loss-function is convex.
%\todo{perhaps here you can add the formula for the influence function  to make it clear what we mean by hessian vector etc product} 
For convex loss functions, influence functions are also accurate even when the model perturbations are fairly large (e.g. in the group influence case \citep{influence2, pmlr-v119-basu20b}). However, when the convexity assumption of the underlying loss function is violated, which is the case in deep learning, the behaviour of influence functions is not well understood and is still an open area of research. With recent advances in computer vision \citep{Szeliski:2010:CVA:1941882}, natural language processing \citep{Sebastiani:2002:MLA:505282.505283}, high-stakes applications such as medicine \citep{medical_imaging}, %and finance \citep{finance}, 
it has become particularly important to interpret deep model predictions. This makes it critical to understand influence functions in the context of deep learning, which is the main focus of our paper.   

%Although \citep{influence1} has developed an influence function in the convex case, it has provided some successful empirical results of the application of the influence function in a non-convex case as well. In particular, \citep{influence1} has shown that for a small CNN with around 2,600 parameters trained on $10 \%$ of MNIST, influence functions are fairly accurate. However, it has remained an open question if a successful application of influence functions can be generalized to different deep model architectures and practical input datasets such as full MNIST, CIFAR-10 and ImageNet. In this paper, we aim to shed some light on this question. 

Despite their non-convexity, it is sometimes believed that influence functions would work for deep networks. The excellent work of \citep{influence1} successfully demonstrated one example of influence estimation for a deep network, a small (2600 parameters), "all-convolutional" network \citep{Springenberg}. To the best of our knowledge, this is the one of the \textit{few} cases for deep networks where influence estimation has been shown to work. A question of key importance to practitioners then arises: for what other classes of deep networks does influence estimation work? In this work, we provide a comprehensive study of this question and find a pessimistic answer: {\it influence estimation is quite fragile for a variety of deep networks.}

In the case of deep networks, several factors might have an impact on influence estimates: (i) due to non-convexity of the loss function, different initializations of the perturbed model can lead to significantly different model parameters (with approximately similar loss values); (ii) even if the initialization of the model is fixed, the curvature values of the network (i.e. eigenvalues of the Hessian matrix) at optimal model parameters might be very large in very deep networks, leading to a substantial Taylor's approximation error of the loss function and thus resulting in poor influence estimates; (iii) for large neural networks, computing the exact inverse-Hessian Vector product, required in computation of influence estimates, can be computationally very expensive. Thus, one needs to use approximate inverse-Hessian Vector product techniques which might be erroneous; resulting in low quality influence estimates; and finally (iv) different architectures can have different loss landscape geometries near the optimal model parameters, leading to varying influence estimates.% across different model architectures.

In this paper, we study aforementioned issues of using influence functions in deep learning through an extensive experimental study on progressively-growing complex models and datasets. We first start our analysis with a case study of a small neural network for the Iris dataset where the exact Hessian matrix can be computed. We then progressively increase the complexity of the network and analyse a CNN architecture (depth of 6) trained on $10 \%$ of MNIST dataset, similar to \citep{influence1}. Next, we evaluate the accuracy of influence estimates for more complex deep architectures (e.g. ResNets) trained on MNIST and CIFAR-10. Finally, we compute influence estimates on the ImageNet dataset using ResNet-50.

%Through our experiments across different architectures and image datasets, we firstly find that influence functions are extremely fragile for deep networks across a wide variety of different settings. Secondly we understand the source of fragility for influence functions in the non-convex setting through a case study of a smaller neural network architecture.
%In such a setting the exact influence function can be computed without any approximation techniques. 
We make the following observations through our analysis:
\begin{itemize}
    \item We find that the network depth and width have a strong impact on influence estimates. In particular, we show that influence estimates are fairly accurate when the network is shallow, while for deeper models, influence estimates are often erroneous. We attribute this partially to the increasing curvature values of the network as the depth increases.
    \item We observe that the weight decay regularization is important to obtain high quality influence estimates in certain architectures and datasets. 
    %\todo{did we have some partial understanding why this is the case?}
    \item We show that the inverse-Hessian Vector product approximation techniques such as stochastic estimation \citep{Agarwal2016SecondOS} are erroneous, especially when the network is deep. This can contribute to the low quality of influence estimates in deep models. 
    \item We observe that the choice of test-point has a substantial impact on the quality of influence estimates, across different datasets and architectures.
    \item In very large-scale datasets such as ImageNet, we have found that even ground-truth influence estimates (obtained by leave-one-out re-training) can be inaccurate and noisy partially due to the model's training and convergence.   
\end{itemize}
These results highlight sensitivity of current influence functions in deep learning and call for developing robust influence estimators to be used in large-scale machine learning applications.
\vspace{-0.6em}
\section{Related Works}
\vspace{-0.6em}
Influence functions are primarily used to identify important training samples for test-time predictions and debug machine learning models \citep{influence1}. Similar to influence functions, \citep{experimental_design} tackles the problem of approximating a dataset using a subset of the dataset. In recent times, there is an increase in the applications of influence functions for tasks other than interpretability. For e.g.\citep{Schulam2019CanYT} has used influence functions to audit the reliability of test-predictions. In NLP, influence functions have been used to detect biases in word-embeddings \citep{influence_word_embeddings} whereas in the domain of ML security, influence functions have been shown to be effective in crafting stronger data-poisoning attacks \citep{koh2019stronger}. Influence functions are also effective in the identification of important training groups (rather than an individual sample) \citep{Basu2019SecondOrderGI, influence2}. Prior theoretical work \citep{Giordano2018ASA, Giordano2019AHS} have focused on quantifying finite sample error-bounds for influence estimates when compared to the ground-truth re-training procedures.  Recently, alternative methods to find influential samples in deep networks have been proposed. In \citep{representer}, test-time predictions are explained by a kernel function evaluated at the training samples. Influential training examples can also be obtained by tracking the change in loss for a test-prediction through model-checkpoints, which are stored during the training time \citep{tracking}. While these alternative methods \citep{representer, tracking} work well for deep networks in interpreting model predictions, they lack the ``jackknife" like ability of influence functions which makes it useful in multiple applications other than interpretability (e.g. uncertainty estimation). 
%\citep{pmlr-v108-barshan20a} introduces a new form of relative influence function to provide local model explanations by using a constraint on the global explanations. However it does not highlight the limitations of the current influence functions in deep learning.
\vspace{-0.7em}
\section{Basics of Influence Function}
\vspace{-0.4em}
Consider $h$ to be a function parameterized by $\btheta$ which maps from an input feature space $\mathcal{X}$ to an output space denoted by $\mathcal{Y}$. The training samples are denoted by the set $\mathcal{S} = \{z_{i} : (x_{i},y_{i}) \}_{i=1}^{n}$, while the loss function is represented by $\ell (h_{\btheta}(z))$ for a particular training example $z$. The standard empirical risk minimization solves the following optimization problem: 
\vspace{-0.4em}
\begin{equation}
    \btheta^{*} = \arg \min_{\btheta} \frac{1}{n} \sum_{i=1}^{n} \ell(h_{\btheta}(z_{i})).
\end{equation}
Up-weighting a training example $z$ by an infinitesimal amount $\epsilon$ leads to a new set of model parameters denoted by  $\btheta_{\{ z \}}^{\eps}$. This set of new model parameters $\btheta_{ \{ z \} }^{\eps}$ is obtained by solving:
\vspace{-0.6em}
\begin{equation}
    \label{upweight_erm}
    \btheta_{ \{ z \} }^{\eps} = \arg \min_{\btheta}  \frac{1}{n} \sum_{i=1}^{n} \ell(h_{\btheta}(z_{i})) + \eps \ell(h_{\btheta}(z)).
\end{equation}
Removing a training point $z$ is similar to up-weighting its corresponding weight by $\epsilon = -1/n$ in Equation(\ref{upweight_erm}). The main idea used by \citep{influence1} is to approximate $\btheta^{\eps}_{\{z\}}$ by the first-order Taylor series expansion around the optimal model parameters represented by $\btheta^{*}$, which leads to:
\begin{equation}
    \label{taylors}
    \btheta^{\eps}_{\{z\}} \approx \btheta^{*} - \eps H_{\btheta^{*}}^{-1} \nabla_{\btheta} \ell(h_{\btheta^{*}}(z)),
\end{equation}
where $H_{\btheta^{*}}$ represents the Hessian with respect to model parameters $\btheta^{*}$. Following the classical result of \citep{cook_influence}, the change in the model parameters ($\Delta\btheta = \btheta^{\eps}_{\{z\}} - \btheta^{*}$) on up-weighting the training example $z$ can be approximated by the influence function ($\mathcal{I}(z)$) as follows: %and is denoted by $\mathcal{I}(z)$ as follows: 
\vspace{-0.4em}
\begin{equation}
    \mathcal{I}(z)= \frac{d \btheta^{\epsilon}_{\{ z\}}}{d \epsilon} |_{\epsilon = 0} = - H_{\btheta^{*}}^{-1} \nabla_{\btheta} \ell\left(h_{\btheta^{*}}(z)\right).
\end{equation}
The change in the loss value for a particular test point $z_{t}$ when a training point $z$ is up-weighted can be approximated as a closed form expression by the chain rule \citep{influence1}: 
\begin{equation}
    \label{inf1}
    \mathcal{I}(z, z_{t})=-\nabla \ell(h_{\btheta^{*}}(z_{t}))^{T}H_{\btheta^{*}}^{-1} \nabla \ell(h_{\btheta^{*}}(z)).
\end{equation}
\vspace{-0.1em}
$\mathcal{I}(z,z_{t})/n$ is approximately the change in the loss for the test-sample $z_{t}$ when a training sample $z$ is removed from the training set. This result is, however, based on the assumption that the underlying loss function is strictly convex in the model parameters $\btheta$ and the Hessian $H_{\btheta^{*}}$ is a positive-definite matrix \citep{influence1}.
%\footnote{For deep networks, where the loss function is non-convex, a small dampening factor $\lambda$ is added to the Hessian $H_{\btheta^{*}}$ to make it positive-definite \citep{influence1}}. 
For large models, inverting the exact Hessian $H_{\btheta^{*}}$ is expensive. In such cases, the inverse-Hessian Vector product can be computed efficiently with a combination of Hessian-vector product \citep{Pearlmutter} and optimization techniques (see Appendix for details).

\begin{figure*}
    \hskip-0.70cm
  \includegraphics[width=15cm, height=3.9cm]{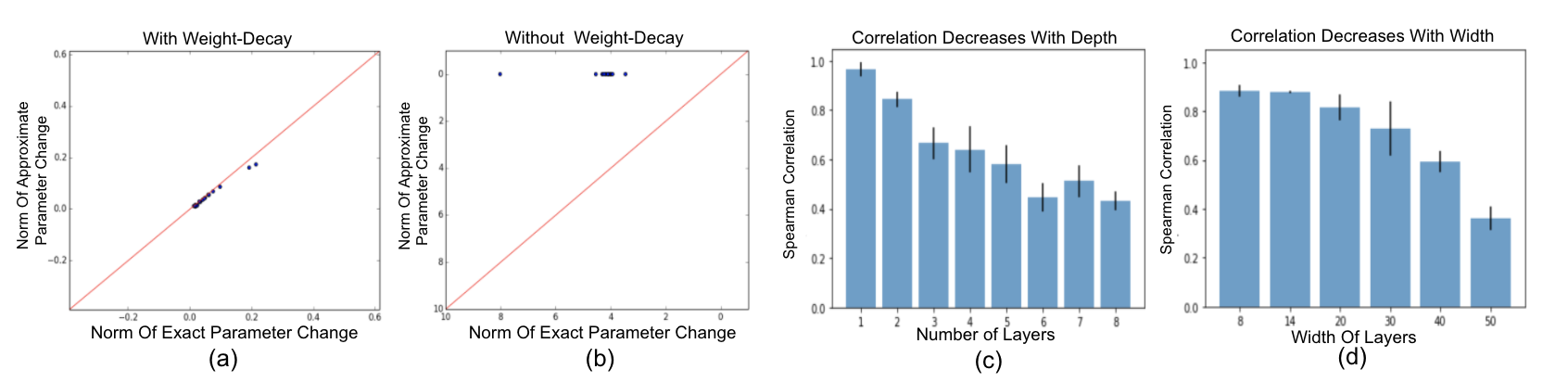}
  \vspace{-2em}
  \caption{\label{figure1} Iris dataset experimental results - (a,b) Comparison of norm of parameter changes computed with influence function vs re-training; (a) trained with weight-decay; (b) trained without weight-decay. (c) Spearman correlation vs. network depth. (d) Spearman correlation vs. network width.}
  \vspace{-0.9em}
\end{figure*}
\vspace{-0.6em}
\section{What Can Go Wrong for Influence Functions In Deep Learning?}
\vspace{-0.3em}
First-order influence functions \citep{influence1} assume that the underlying loss function is convex and the change in model parameters is small when the empirical weight distribution of the training data is infinitesimally perturbed. In essence, this denotes the Taylor's gap in Equation (\ref{taylors}) to be small for an accurate influence estimate. However in the case of non-convex loss functions, this assumption is \textit{not} generally true. Empirically, we find that the Taylor's gap is strongly affected by common hyper-parameters for deep networks. For example, in Fig. (\ref{figure1})-(a,b), we find that for networks trained without a weight-decay regularization on Iris, the Taylor's gap is large resulting in low quality influence estimates. In a similar vein, when the network depth and width is considerably large (i.e. the over-parameterized regime), the Taylor's gap increases and substantially degrades the quality of influence estimates (Fig. (\ref{figure2})). Empirically this increase in Taylor's gap strongly correlates with the curvature values of the loss function evaluated at the optimal model parameters as observed in Fig. (\ref{figure2}-(b)). 

Further complications may arise for larger models, where influence estimations in such settings require an additional approximation to compute the inverse-Hessian vector product. Nonetheless, we observe in Fig. (\ref{figure2})-(a), that on Iris this approximation has only a marginal impact on the influence estimation. These results show that that network architecture, hyper-parameters, and loss curvatures are important factors for proper influence estimations. 
%For very large scale datasets such as ImageNet, not only the influence estimates but the ground-truth influence estimation can also be noisy thus hindering the ability of . 
In the next section, we discuss these issues in details through controlled experiments on datasets and models of increasing complexity.
\vspace{-1em}
\section{Experiments}
\vspace{-0.6em}
\textbf{Datasets}: We first study the behaviour of influence functions in a small Iris dataset \citep{iris}, where the exact Hessian can be computed. Further, we progressively increase the complexity of the model and datasets: we use small MNIST \citep{influence1} to evaluate the accuracy of influence functions in a small CNN architecture with a depth of 6. Next, we study influence functions on modern deep architectures trained on the standard MNIST \citep{mnist} and CIFAR-10 \citep{cif} datasets. Finally, to understand how influence functions scale to large datasets, we use ImageNet \citep{imagenet_cvpr09} to compute the influence estimates. 

\textbf{Evaluation Metrics}: We evaluate the accuracy of influence estimates at a given test point $z_{t}$ using both Pearson \citep{pearson} and Spearman rank-order correlation \citep{spearman04} with the ground-truth (obtained by re-training the model) across a set of training points. Most of the existing interpretability methods desire that influential examples are ranked in the correct order of their importance \citep{Ghorbani2017InterpretationON}. Therefore, to evaluate the accuracy of influence estimates, Spearman correlation is often a better choice. 
%In our experiments in particular, we evaluate the correlations with the top influential training points. 
%We provide more details on them in the subsequent sections. 
\vspace{-0.6em}
\subsection{Understanding Influence Functions when the Exact Hessian Can be Computed}
\vspace{-0.3em}
\label{exact}
\textbf{Setup:} Computing influence estimates with the exact Hessian has certain advantages in our study: a) it bypasses inverse-Hessian Vector product approximation techniques which induce errors in computing influence estimates. Thus, we can compare influence estimates computed with exact vs. approximate inverse-Hessian Vector products to quantify this type of error; b) The deviation of the parameters computed with the influence function from the exact parameters can be computed exactly. This information can be useful to further quantify the error incurred by (first-order) influence estimates in the non-convex setup. However, computations of the exact Hessian matrix and its inverse are only computationally feasible for models with small number of parameters. Thus, we use the Iris dataset along with a small feed-forward neural network to analyse the behaviour of influence function computed with the exact Hessian in a non-convex setup. We train models to convergence for 60k iterations with full-batch gradient descent. To obtain the ground-truth estimates, we re-train the models for 7.5k steps, starting from the optimal model parameters. For our analysis, we choose the test-point with the maximum loss and evaluate the accuracy of influence estimates with the ground-truth amongst of the top $16.6 \%$ of the training points. Through our experiments with the exact Hessian, we answer some relevant questions related to how properties of the network such as depth, width and regularizers (e.g. weight-decay) affect the influence estimates.
\begin{figure}[t]
    \hskip-0.9cm
  \includegraphics[width=16cm]{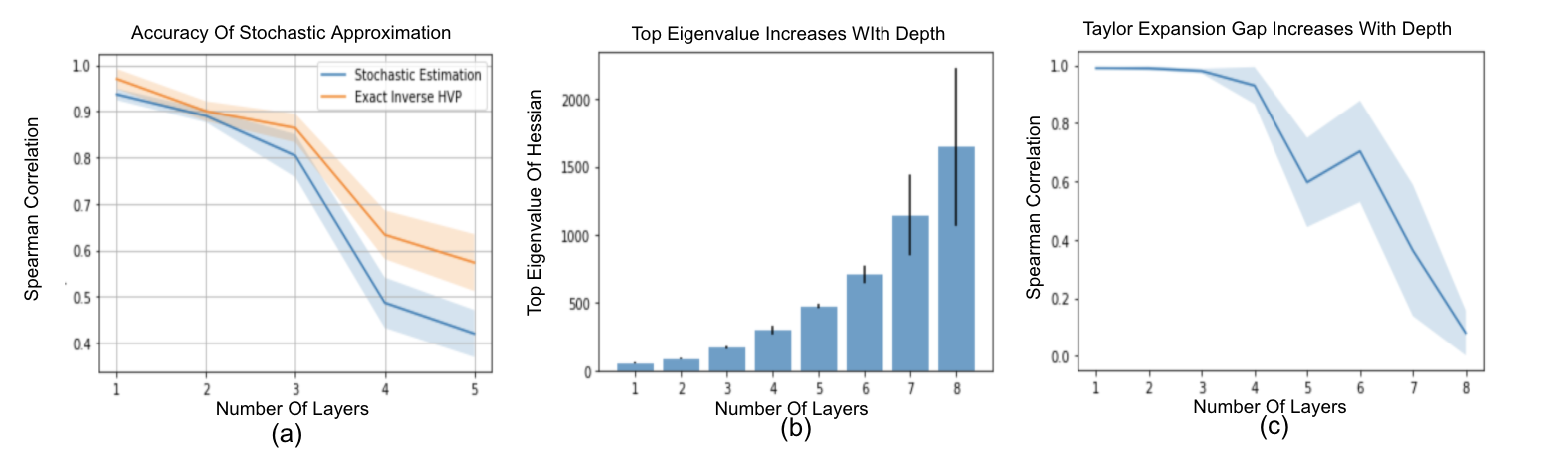}
  \vspace{-1.2em}
  \caption{\label{figure2} Iris dataset experimental results; (a) Spearman correlation of influence estimates with the ground-truth estimates computed with stochastic estimation vs. exact inverse-Hessian vector product. (b) Top eigenvalue of the Hessian vs. the network depth. (c) Spearman correlation between the norm of parameter changes computed with influence function vs. re-training. }
\vspace{-1.3em}
\end{figure} 

\textbf{The Effect of Weight-Decay}: 
One of the simpler and common regularization techniques used to train neural networks is weight-decay regularization. In particular, a term $\lambda \| \btheta \|_{2}^{2}$, penalizing the scaled norm of the model parameters is added to the objective function, during training, where $\lambda$ is a hyperparameter which needs to be tuned. We train a simple feed-forward network\footnote{With width of 5, depth of 1 and ReLU activations} with and without weight-decay regularization. For the network trained with weight-decay, we observe a Spearman correlation of 0.97 between the influence estimates and the ground-truth estimates. In comparison, for the network trained without a weight-decay regularization, the Spearman correlation estimates decrease to 0.508. In this case, we notice that the Hessian matrix is singular, thus a damping factor of 0.001 is added to the Hessian matrix, to make it invertible. To further understand the reason for this decrease in the quality of influence estimates, we compare the following metric across all training examples: a) Norm of the model parameter changes computed by re-training; b) Norm of the model parameter changes computed using the influence function (i.e. $ \| H_{\btheta^{*}}^{-1} \nabla \ell(z_{i}) \|_{2} \hspace{0.3cm} \forall i \in [1,n]$) (Fig. \ref{figure1}-(a,b)). We observe that when the network is trained without weight-decay, changes in model parameters computed with the influence function have a substantially larger deviation from those computed using re-training. This essentially suggests that the gap in Taylor expansion, using (first-order) influence estimates is large, when the model is trained without weight-decay. We observe similar results with smooth activation functions such as tanh (see the Appendix for details). 

\textbf{The Effect Of Network Depth}:  From Fig. \ref{figure1}-(c), we see that network depth has a dramatic effect on the quality of influence estimates. For example, when the depth of the network is increased to 8, we notice a considerable decrease in the Spearman correlation estimates. To further our understanding about the decrease in the quality of influence estimates when the network is deeper, we compute the gap in the approximation between the ground-truth parameter changes (computed by re-training) and the approximate parameter changes (computed using the influence function). To quantify the error gap, we compute the Spearman correlation estimates between the norm of true and approximate parameter changes across the top $16.6 \%$ of the influential examples.
We find that with increasing depth, the Spearman correlation estimates between the norm of the true and approximate parameter changes decrease. From Fig. \ref{figure2}-(c), we see that the approximation error gap is particularly large when the depth of the network is more than 5. We also notice a consistent increase in the curvature of the loss function (Fig. \ref{figure2}-(b)), as the network becomes deeper. This possibly suggests that the curvature information of the network can be an upper bound in the approximation error gap between the true parameters and the ones computed using the influence function. Even in case of non-smooth activation functions like ReLU, we have a similar observation. (see the Appendix for more details). 
\begin{figure}[t!]
    \hskip-0.62cm
  \includegraphics[width=15cm, height=4.7cm]{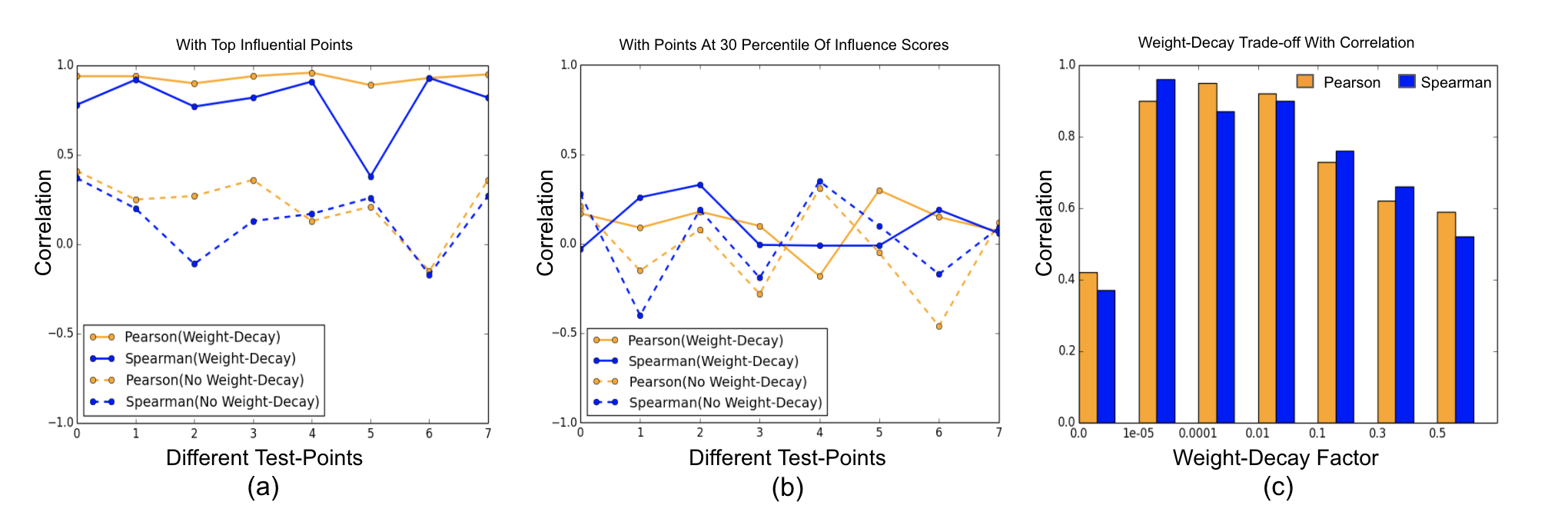}
  \vspace{-2em}
  \caption{\label{small_influence} Experiments on small MNIST using a CNN architecture. (a) Estimation of influence function  with and without weight decay on (a) the top influential points, (b) training points at $30^{th}$ percentile of influence score distribution. (c) Correlation vs the weight decay factor (evaluated on the top influential points).}
  \vspace{-1.5em}
\end{figure} 

\textbf{The Effect Of Network Width}: 
To see the effect of the network width on the quality of influence estimates, we evaluate the influence estimates for a feed-forward network of constant depth, by progressively increasing its width. From Fig. \ref{figure1}-(d), we observe that with an increase in network width, the Spearman correlation decreases consistently. For example, we find that the Spearman correlation decreases from 0.82 to 0.56, when the width of the network is increased from 8 to 50. This observation suggests that over-parameterizing a network by increasing its width has a strong impact in the quality of influence estimates. 

\textbf{The Effect of Stochastic Estimation on inverse-Hessian Vector Product}:
For large deep networks, the inverse-Hessian Vector product is computed using stochastic estimation\citep{Agarwal2016SecondOS}, as the exact Hessian matrix cannot be computed and inverted. To understand the effectiveness of stochastic approximation, we compute the influence estimates with both the exact Hessian and stochastic estimation.  We observe that across different network depths, the influence estimates computed with stochastic estimation have a marginally lower Spearman correlation when compared to the ones computed with the exact Hessian. From Fig. \ref{figure2}-(a), we find that the error in the approximation is more, when the network is deeper. 
\vspace{-0.6em}
\subsection{Understanding Influence Functions in Shallow CNN Architectures}
\vspace{-0.4em}
\textbf{Setup}: In this section, we perform a case study using a CNN architecture\footnote{The model has 2600 parameters and is trained for 500k iterations to reach convergence with the optimal model parameters $\btheta^{*}$. The ground-truth estimates are obtained by re-training the models from the optimal parameter set $\btheta^{*}$ for 30k iterations. When trained with a weight-decay, a regularization factor of 0.001 is used.} on the small MNIST dataset (i.e. $10 \%$ of MNIST); a similar setup used in \citep{influence1}. To assess the accuracy of influence estimates, we select a set of test-points with high test-losses computed at the optimal model parameters. For each of the test points, we select 100 training samples with the highest influence scores and compute the ground-truth influence by re-training the model. We also select 100 training points with influence scores at the $30^{th}$ percentile of the entire influence score distribution. These training points have low influence scores and a lower variance in their scores when compared to the top influential points. The model is trained with and without weight-decay regularization. 

When trained with a weight-decay and evaluated based on the top influential points, we find that the correlation estimates are consistently significant (Fig. \ref{small_influence}-(a)). This is consistent with the results reported in \citep{influence1}. However, when the evaluation is done with the set of training samples at the $30^{th}$ percentile of the influence score distribution, the correlation estimates decrease significantly (Fig. \ref{small_influence}-(b)). This shows that influence estimates of only the top influential points are precise when compared to ground-truth re-trainings. Furthermore, without the weight-decay regularization, influence estimates in both cases are poor across all the test-points (Fig. \ref{small_influence}-(a,b)). 

To further understand the impact of weight-decay on influence estimates, we train the network with different weight-decay regularization factors. From Fig. \ref{small_influence}-(c), we see that the selection of weight-decay factor is important in getting high-quality influence estimates. For this specific CNN architecture, we notice that the correlations start decreasing when the weight-decay factor is greater than 0.01. Moreover, from Fig. \ref{small_influence}-(a,b), we find that the selection of test-point also has a strong impact on the quality of influence estimates. For example, when the network is trained with weight-decay and the influence estimates are computed for top influential training points, we notice that the Spearman correlation estimates range from 0.92 to 0.38 across different test-points and have a high variance.  

These results show that despite some successful applications of influence functions in this non-convex setup, as reported in \citep{influence1}, their performances are very sensitive to hyper-parameters of the experiment as well as to the training procedure. In the next two sections, we assess the quality of influence estimates on more complex architectures and datasets including MNIST, CIFAR-10 and ImageNet. In particular, we desire to understand, if the insights gained from experiments on smaller networks can be generalized to more complex networks and datasets.
\vspace{-0.6em}
\subsection{Understanding Influence Functions in Deep Architectures}
\textbf{Setup}: In this section, we evaluate the accuracy of influence estimates using MNIST and CIFAR-10 datasets across different network architectures including small CNN\citep{influence1}, LeNet \citep{lenet}, ResNets \citep{res_nets}, and VGGNets \citep{vggnetwork}\footnote{For CIFAR-10, evaluations on small CNN have not been performed due to the poor test accuracy.}. To compute influence estimates, we choose two test points for each architecture: a) the test-point with the highest loss, and b) the test-point at the $50^{th}$ percentile of the losses of all test points. For each of these two test points, we select the top 40 influential training samples and compute the correlation of their influence estimates with the ground-truth estimates. To compute the ground-truth influence estimates, we follow the strategy of \citep{influence1}, where we re-train the models from optimal parameters for $6\%$ of the steps used for training the optimal model. When the networks are trained with a weight-decay regularization, we use a constant weight-decay factor of 0.001 across all the architectures (see Appendix for more details). 
\begin{table}
\resizebox{\columnwidth}{!}{
\begin{tabular}{|c|c|c|c|c|c|c|c|c|c|c|c|c|}
\hline
\textbf{Dataset} & \multicolumn{6}{c|}{\textbf{MNIST}} & \multicolumn{6}{c|}{\textbf{CIFAR-10}} \\ \hline
 & \multicolumn{2}{c|}{\textbf{\begin{tabular}[c]{@{}l@{}}A\\ (With \\ Decay)\end{tabular}}} & \multicolumn{2}{c|}{\textbf{\begin{tabular}[c]{@{}l@{}}B\\ (With \\ Decay)\end{tabular}}} & \multicolumn{2}{c|}{\textbf{\begin{tabular}[c]{@{}l@{}}A\\ (Without \\ Decay)\end{tabular}}} & \multicolumn{2}{c|}{\textbf{\begin{tabular}[c]{@{}l@{}}A \\ (With \\ Decay)\end{tabular}}} & \multicolumn{2}{c|}{\textbf{\begin{tabular}[c]{@{}l@{}}B\\ (With \\ Decay)\end{tabular}}} & \multicolumn{2}{c|}{\textbf{\begin{tabular}[c]{@{}l@{}}A\\ (Without \\ Decay)\end{tabular}}} \\ \hline
\textbf{Architecture} & P & S & P & S & P & S & P & S & P & S & P & S \\ \hline
Small CNN & \textbf{0.95} & \textbf{0.87} & \textbf{0.92} & \textbf{0.82} & \textbf{0.41} & 0.35 & - & - & - & - & - & - \\ \hline
LeNet & \textbf{0.83} & \textbf{0.51} & 0.28 & 0.29 & 0.18 & 0.12 & \textbf{0.81} & \textbf{0.69} & 0.45 & 0.46 & 0.19 & 0.09 \\ \hline
VGG13 & 0.34 & \textbf{0.44} & 0.29 & 0.18 & 0.38 & 0.31 & \textbf{0.67} & \textbf{0.63} & \textbf{0.66} & \textbf{0.63} & \textbf{0.79} & \textbf{0.73} \\ \hline
VGG14 & 0.32 & 0.26 & 0.28 & 0.22 & 0.21 & 0.11 & \textbf{0.61} & \textbf{0.59} & \textbf{0.49} & 0.41 & \textbf{0.75} & \textbf{0.64} \\ \hline
ResNet18 & \textbf{0.49} & 0.26 & 0.39 & 0.35 & 0.14 & 0.11 & \textbf{0.64} & \textbf{0.42} & 0.25 & 0.26 & \textbf{0.72} & \textbf{0.69} \\ \hline
ResNet50 & 0.24 & 0.22 & 0.29 & 0.19 & 0.08 & 0.13 & \textbf{0.46} & 0.36 & 0.24 & 0.09 & 0.32 & 0.14 \\ \hline
\end{tabular} } \\ \\
\vspace{-1.5em}
\caption{\label{influence_deep}Correlation estimates on MNIST And CIFAR-10 ; A=Test-point with highest loss; B=Test-point at the $50^{th}$ percentile of test-loss spectrum; P=Pearson correlation; S=Spearman correlation }
\vspace{-1.2em}
\end{table}
 
\textbf{Results On MNIST}: From Table \ref{influence_deep}, we observe that for the test-point with the highest loss, the influence estimates in the small CNN and LeNet architectures (trained with the weight-decay regularization) have high qualities. These networks have $2.6$k and $44$k parameters, respectively, and are relatively smaller and less deep than the other networks used in our experimental setup. As the depth of the network increases, we observe a consistent decrease in quality of influence estimates. For the test-point with a loss at the $50^{th}$ percentile of test-point losses, we observe that influence estimates {\it only} in the small CNN architecture have good qualities.    

\textbf{Results On CIFAR-10}: 
For CIFAR-10, across all architectures trained with the weight-decay regularization, we observe that the correlation estimates for the test-point with the highest loss are highly significant. For example, the correlation estimates are above 0.6 for a majority of the network architectures. However, for the test-point evaluated at the $50^{th}$ percentile of the loss, the correlations decrease marginally across most of the architectures. We find that on CIFAR-10, even architectures trained without weight-decay regularization have highly significant correlation estimates when evaluated with the test-point which incurs the highest loss.% \todo{we should have a discussion here, very important to somehow explain what is happening, because this is somehow counter-intuitive}   
%\todo{Highlighted for @Soheil to check}

In case of MNIST, we have found that in shallow networks, the influence estimates are fairly accurate while for deeper networks, the quality of influence estimates decrease. For CIFAR-10, although the influence estimates are significant, we found that the correlations are marginally lower in deeper networks such as ResNet-50. The improved quality of influence estimates in CIFAR-10 can be attributed to the fact that for a similar depth, architectures trained on CIFAR-10 are less over-parameterized compared to architectures trained on MNIST. Note that, in Section \ref{exact}, where the exact Hessian matrix can be computed, we observed that over-parameterization decreases the quality of influence estimates. From Table(\ref{influence_deep}), we also observed that the selection of test-point has a sizeable impact on the quality of influence estimates. Furthermore, we noticed large variations in the quality of influence estimates across different architectures. In general we found that influence estimates for small CNN and LeNet are reasonably accurate, while for ResNet-50, the quality of estimates decrease across both MNIST and CIFAR-10. Precise reasons for these variations are difficult to establish. We hypothesize that it can be due to the following factors: (i) Different architectures trained on different datasets have contrasting characteristics of loss landscapes at the optimal parameters which can have an impact on influence estimates. (ii) The weight-decay factor may need to be set differently in various architectures, to obtain high quality influence estimates.  \\ \\
\begin{wrapfigure}{r}{0.45\textwidth}
\vspace{-10pt}
\hskip-1em
    \includegraphics[width=0.45\textwidth]{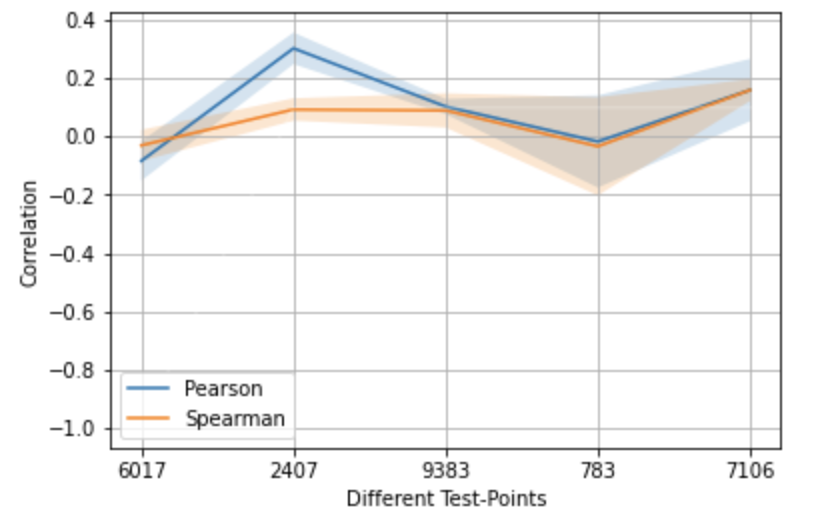}
    \vspace{-10pt}
    \label{cifar100}
  \caption{\label{cifar_100}Influence for CIFAR-100}
  \vspace{-20pt}
\end{wrapfigure}
\textbf{Results on CIFAR-100}: In the case of CIFAR-100, we train a ResNet-18 model with a weight-decay regularization factor of $5e^{-4}$. The influence estimates are then computed for test-points with the highest losses (Index: 6017, 2407, 9383) and test-points around the $50^{th}$ percentile of the test loss (Index: 783, 7106) over multiple model initialisations. Unlike in the case of MNIST and CIFAR-10, from Fig. \ref{cifar_100} we observe the correlation estimates to be of substantially poor quality. We provide additional visualizations of the influential training examples in the Appendix section.
\vspace{0.2em}
\subsection{Is Scaling Influence Estimates To ImageNet Possible?}
\vspace{-0.4em}
The application of influence functions to ImageNet scale models provides an appealing yet challenging opportunity. It is appealing because, if successful, it opens a range of applications to large-scale image models, including interpretability, robustness, data poisoning, and uncertainty estimation. It is challenging for a number of reasons. Notable among these is the high computational cost of training and re-training, which limits the number of ground truth evaluations. In addition, all of the previously discussed difficulties in influence estimations still remain, including (i) non-convexity of the loss, (ii) selection of scaling and damping hyperparameters in the stochastic estimation of the Hessian, and (iii) the lack of convergence of the model parameters.  The scale of ImageNet raises additional questions about the feasibility of leave-one-out retraining as the ground truth estimator. Given that there are 1.2M images in the training set, \textit{is it even possible that the removal of one image can significantly alter the model?} In other words, we question whether or not reliable ground truth estimates may be obtained through leave-one-out re-training at this scale.

To illustrate this, we conduct an additional influence estimation on ImageNet. After training an initial model to 92.302\% top5 test accuracy, we select two test points at random, calculate influence over the entire training set, and then select the top 50 points by their influences as candidates for re-training. We then use the re-training procedure suggested by \citep{influence1}, which starts leave-one-out re-training from the parameter set obtained after the initial training. We re-train for an additional 2 epochs, approximately 5\% of the original training time, and calculate the correlations. We observe that for both test points, both Pearson and Spearman correlations are very low (less than $0.15$, see details in the Appendix).  

%For test point 13,923, we find the Pearson and Spearman correlations to both be 0.15. For test point 2257, we find the Pearson and Spearman correlations to be 0.03 and 0.12 respectively. Additional experimental details are included in the Supplementary.
%The correlations above are weak, showing that the influence calculations have broken down.  

In our experiments, we observe high variability among ground-truth estimates obtained by re-training the model (see the appendix for details). We conjecture that this may be partially due to the fact that the original model has not be fully converged. To study this, we train the original model with \textit{all} training points for an additional 2 epochs and measure the change in the test loss. We find that the overall top5 test accuracy has improved slightly to 92.336 \% (+0.034) and the loss for one of the considered test points has decreased by relatively a significant amount of 0.679. However, the loss for the other point has increased slightly by 0.066. Such changes in loss values can therefore out-power the effect of leave-one-out re-training procedure. Second, we calculate the $2$-norm of the weight gradients, which should be close to zero near an optimal point, and compare it to a standard pre-trained ImageNet ResNet-50 model as a baseline. We find these norms to be 20.18 and and 15.89, respectively, showing our model has similar weight gradient norm to the baseline. Although these norms are relatively small given that there are 25.5M parameters, further re-training the model still changes loss values for some samples considerably, making the ground-truth estimates noisy. We suggest that one way to obtain reliable ground-truth influence estimates in such large models can be through assessing the influence of a group of samples, rather than a single one.
\vspace{-0.8em}
\section{Discussion on ground-truth influence}
In our experimental setup, to obtain the ground-truth influence, we follow the strategy of re-training from optimal model parameters as shown in \citep{influence1, influence2}. Even for moderately sized datasets and architectures, re-training from scratch (instead of re-training from optimal model parameters) is computationally expensive. Although re-training from optimal model parameters is an approximation compared to re-training from scratch, we notice that the approximation works quite well in practice. To validate the effectiveness of this strategy, we first compute the norm of the difference in parameters obtained by re-training from scratch vs. re-training from optimal parameters. Next we compute the correlation between the influence estimates and ground-truth using both the re-training strategies. 
\begin{figure}[t!]
    \hskip-0.2cm
  \includegraphics[width=15cm, height=4.7cm]{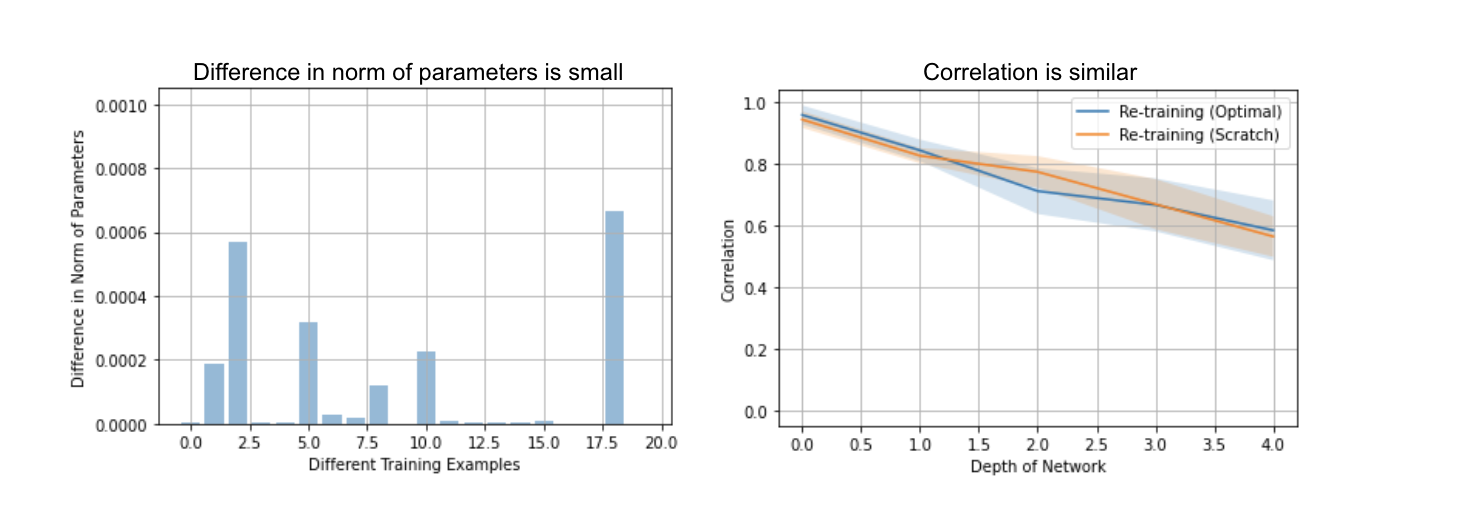}
  \vspace{-2em}
  \caption{\label{retrained} (a) Difference in norm of parameters obtained by re-training from scratch vs. re-training from optimal parameters. (b) Correlation estimates with re-training from scratch vs. re-training from optimal parameters.}
  \vspace{-1em}
\end{figure} 
From Fig. \ref{retrained}, we observe the norm of parameter differences using the two re-training strategies to be small. Similarly using both the re-training strategies as ground-truth yield similar correlation estimates. These results highlight that re-training from optimal parameters (although an approximation) is close to re-training from scratch.

\section{Conclusion}
\vspace{-0.7em}
In this paper, we present a comprehensive analysis of the successes and failures of influence functions in deep learning. Through our experiments on datasets including Iris, MNIST, CIFAR-10, CIFAR-100, ImageNet and architectures including LeNet, VGGNets, ResNets, we have demonstrated that influence functions in deep learning are fragile in general. We have shown that several factors such as the weight-decay, depth and width of the network, the network architecture, stochastic approximation and the selection of test points, all have strong effects in the quality of influence estimates. In general, we have observed that influence estimates are fairly accurate in shallow architectures such as small CNN\citep{influence1} and LeNet, while in very deep and wide architectures such as ResNet-50, the estimates are often erroneous. Additionally, we have scaled up influence computations to the ImageNet scale, where we have observed influence estimates are highly imprecise. %when compared to the ground-truth re-trainings. 
These results call for developing robust influence estimators in the non-convex setups of deep learning. 
\section{Acknowledgements}
Authors thank Daniel Hsu, Alexander D’Amour and Pang Wei Koh for helpful discussions. This
project was supported in part by NSF CAREER AWARD 1942230, HR001119S0026-GARD-FP-052,
AWS Machine Learning Research Award, a sponsorship from Capital One, and Simons Fellowship
on “Foundations of Deep Learning".

\bibliography{iclr2021_conference}
\bibliographystyle{iclr2021_conference}

\newpage 

\section{Appendix}

\subsection{Additional Experimental Results on Iris Dataset}
\begin{figure}[!t]
  \hskip-0.9cm
  \includegraphics[width=16cm]{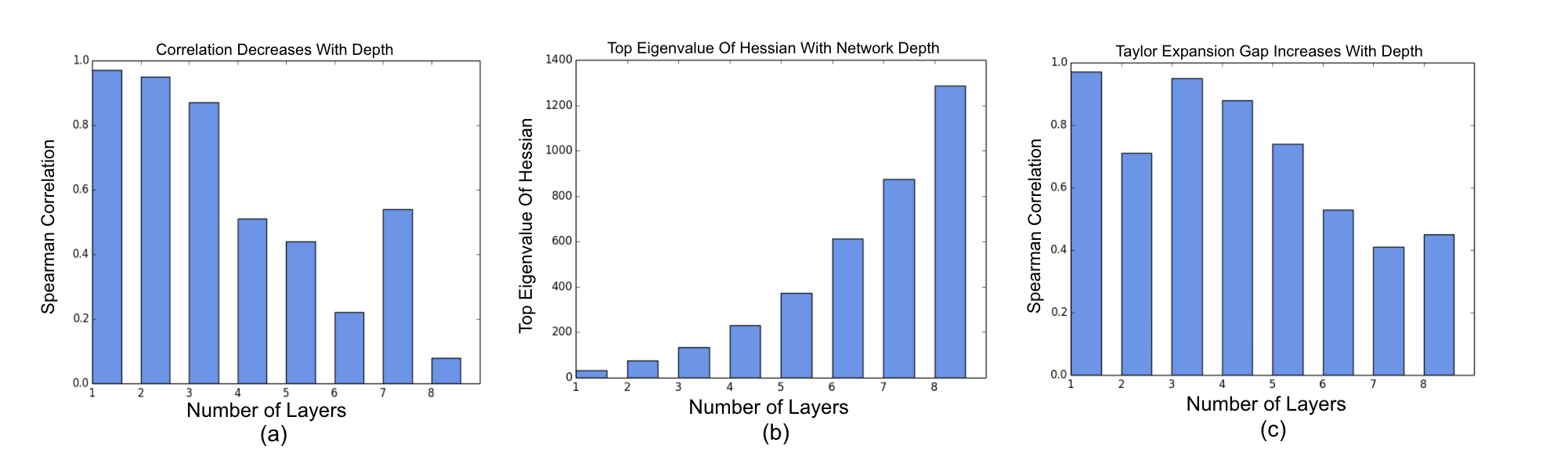}
  \caption{\label{relu-depth} Additional Iris experimental results for ReLU networks: (a) Spearman correlation vs. network depth; (b) Top eigenvalue of the Hessian vs.network depth; (c) Spearman correlation between the norm of parameter changes computed with influence function vs. re-training.  }
\end{figure}

\begin{figure}[ht]
  \includegraphics[width=14cm]{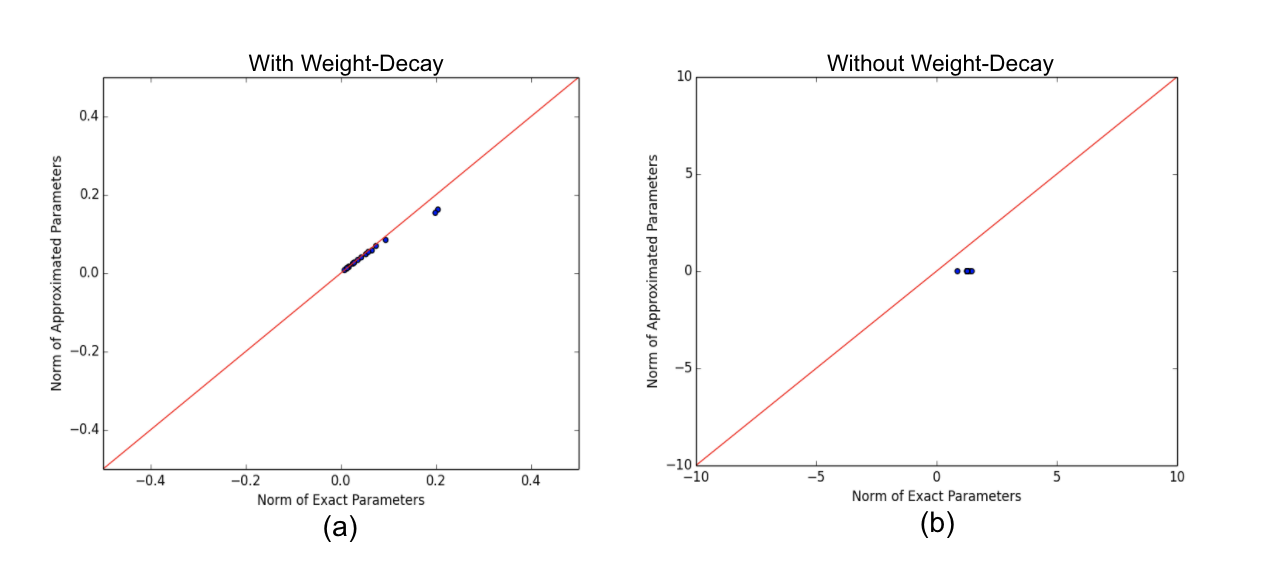}
  \caption{\label{tanh-decay} Additional Iris experimental results for tanh networks; (a) When trained with weight-decay, the Taylor's approximation gap is small; (b) When trained without weight-decay, the Taylor's expansion gap is large. These results are similar to our findings for ReLU networks which are reported in the main text.}
\end{figure}

%\myred{say a few words what additional experimental results we are providing here. also say that these results are consistent with XXX in the main text. }
In this section, we provide additional experimental results to understand the effect of network depth on the correlation estimates for ReLU networks. From Fig. \ref{relu-depth}, we observe that even in case of architectures trained with non-smooth activation functions such as ReLU, the correlation estimates consistently decrease with depth. Similar to our findings in case of networks trained with tanh activation (as shown in the main text), we observe that the top eigenvalue of the Hessian matrix and the Taylor's approximation gap increases with depth.
In the main text, we reported that when a network with ReLU activation is trained with a weight-decay regularization, the correlation estimates are significant and the Taylor's approximation gap is less. We find a similar result even with smoother activation functions such as tanh. From Fig. \ref{tanh-decay}, we observe that when a network with tanh activation is trained with a weight-decay regularization, the Taylor's approximation gap is less. However when the network is trained without a weight-decay regularization, the Taylor's expansion gap is large resulting in poor quality of influence estimates. 
\subsection{What does weight-decay do?}
In our experiments, we observe that with increasing network depth, the correlation between the influence estimates and the ground-truth estimates decrease considerably. 
\begin{wrapfigure}{r}{0.42\textwidth}
\vspace{-1em}
\hskip-0em
  \begin{center}
    \includegraphics[width=0.42\textwidth]{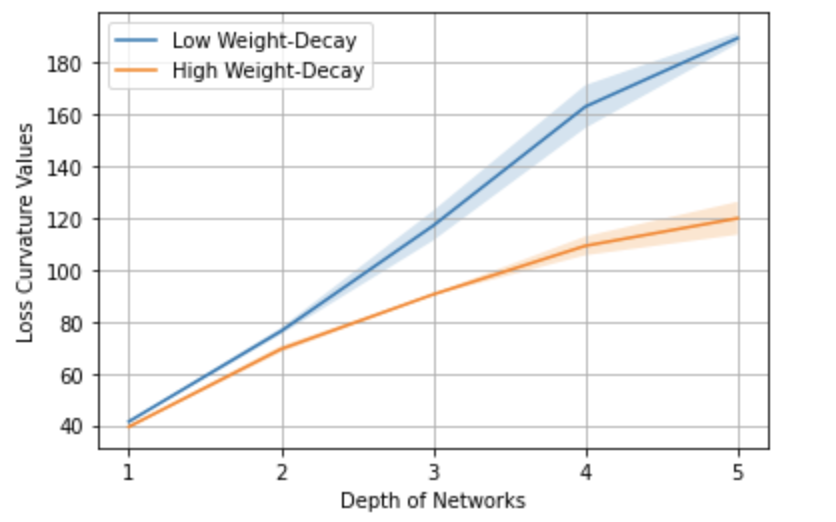}
  \end{center}
  \vspace{-1.5em}
  \caption{\label{decay_loss}Correlations with different training samples}
\end{wrapfigure}
Additionally with increasing depth, the loss curvature values increase. We notice that with a high-value of weight-decay, the loss curvature for deeper networks decrease, which also leads to improvement in correlation values between the influence estimates and the ground-truth. For e.g. in Fig. \ref{decay_loss}, with a weight-decay value of 0.03, the Spearman correlation estimates are 0.47. With a relatively higher weight-decay factor of 0.075, the correlation values improve to 0.72. Increasing the weight-decay factor from 0.03 to 0.075, also decreases the loss curvature values substantially. These results highlight that the selection of weight-decay factor is crucial to obtain high-quality influence estimates, especially for deeper overparameterized networks. 

\subsection{Visualisation of Top Influential Points}
\begin{figure*}[ht]
  \includegraphics[width=14cm]{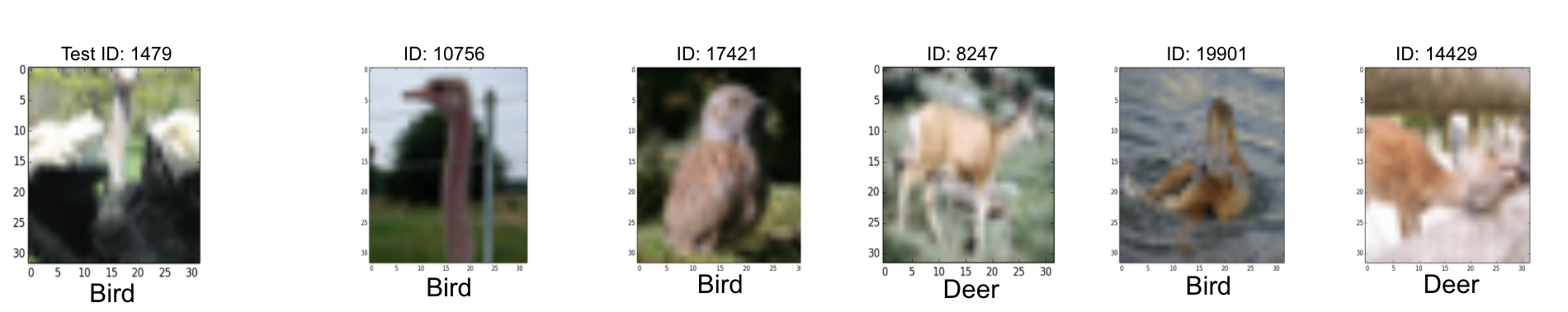}
  \caption{\label{cifar_1}Top 5 influential points for the test point: 1479 (CIFAR-10). The model is a ResNet-18 trained with a weight-decay regularization; Only 3 out of the 5 points are semantically similar to the test-point with class "Bird".}
\end{figure*}

\begin{figure}[H]
  \includegraphics[width=14cm]{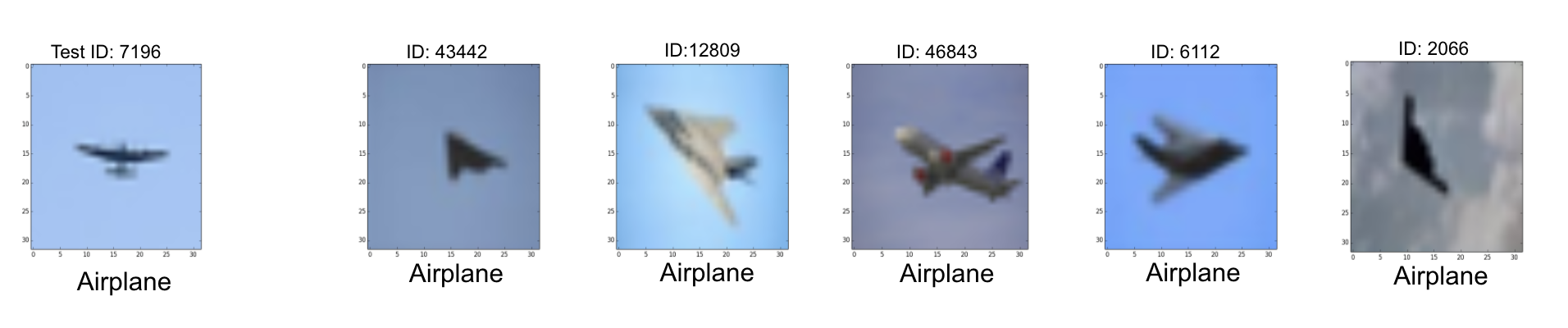}
  \caption{\label{cifar_2}Top 5 influential points for the test point: 7196 (CIFAR-10). The model is a ResNet-18 trained with a weight-decay regularization; All the 5 training points are semantically similar to the test-point from the class "Airplane".}
\end{figure}

In this section, we visualise the top influential training samples corresponding to a given test-point. In the main text, we noted that the selection of test-points has a strong impact on the quality of influence estimates. Additionally, we also observe that the selection of test-points has an impact on the semantic-level similarities between inferred influential training points and the test-points being evaluated. For example, in Fig. \ref{cifar_1}, we observe that 2 out of the top 5 influential points are not from the same class as the test-point with index 1479. However in Fig. \ref{cifar_2}, we observe that all the top 5 influential training samples are semantically similar and from the same class as the evaluated test-point with index 7196. 

\subsection{Running Times}
In this section, we provide computational running times for (first-order) influence function estimations. We note that in models with a large number of parameters, the influence computation is relatively slow. However, even in large deep models, it is still faster than re-training the model for every training example. In our implementation, for a given test-point $z_{test}$, we first compute $c = H_{\btheta^{*}}^{-1}\nabla \ell(h_{\btheta^{*}}(z_{test}))$ once which is the most computationally expensive step. We then compute a vector dot product i.e. $c^{T}\nabla \ell(h_{\btheta^{*}}(z_{i})) \hspace{0.2em} \forall i \in [1,n]$. In Table \ref{times}, we provide the computational running times for estimating influence functions in different network architectures. 

\begin{table}

\resizebox{\columnwidth }{!}{
\begin{tabular}{|c|c|c|}
\hline
\textbf{Architecture} & \textbf{\begin{tabular}[c]{@{}c@{}}Influence Computation Time \\  (MNIST)\end{tabular}} & \textbf{\begin{tabular}[c]{@{}c@{}}Influence Computation Time\\  (CIFAR-10)\end{tabular}} \\ \hline
Small CNN & $141.13 \pm 0.51$ &  N/A \\ \hline
LeNet & $162.6 \pm 2.20 $ & $136.39 \pm 3.16$\\ \hline
VGG13 & $3886.23 \pm 3.45 $  & $4416.54 \pm 2.01 $ \\ \hline
VGG14 & $4619.11 \pm 5.08$ &  $4620.69 \pm 6.11 $  \\ \hline
ResNet-18 & $960.08 \pm 4.67 $  &  $910.58 \pm 8.49 $\\ \hline
ResNet-50 & $4323.13 \pm 8.26$ & $3857.66 \pm 21.6$ \\ \hline
\end{tabular}} \\ \\
\caption{\label{times} Computational running times for influence function across different architectures}
\end{table}

\subsection{Additional Experimental Details on ImageNet Influence Calculations}

In this section we give further details on the influence estimation on ImageNet. To help address the high computational cost of training and re-training, we utilize highly optimized ImageNet training schemes such as those submitted to the DAWNBench competition \citep{dawnbench}. In particular we use the scheme published from \citep{howard2018fastai}\footnote{https://www.fast.ai/2018/08/10/fastai-diu-imagenet/}, for the ResNet-50 architecture which uses several training tricks including progressive image resizing, weight decay tuning, dynamic batch sizes \citep{GoyalDGNWKTJH17}, learning rates \citep{smith}, and half-precision floats. Although these techniques are unorthodox, they are sufficient for our purposes since we need only to compare between the fully trained and re-trained models. We replicate this scheme and obtain a top-5 validation accuracy of 92.302\%. 

We now give further details on the test points selected. The first has a test loss at the 83rd percentile (loss=2.634, index = 13,923, class=kit fox), the second has the test loss at the 37th percentile (loss$=$0.081, index = 2,257, class $=$gila monster), where the indices refer to where they appear in
{\fontfamily{qcr}\selectfont
test\_loader.loader.dataset 
}. We visualize these test points in Figure \ref{imagenet-test}.

Next, for each of these test points, we compute influence across the entire dataset and select the top 50 \textit{training} points by influence scores. We visualize 25 of these points in Figures \ref{13293-train} and \ref{2257-train}. We observe that there is qualitative similarity between the test points and \textit{some} of their respective most influential training points, but not others. Although there is qualitative similarity is some cases, the results are still overall weak quantitatively

\begin{figure*}[ht]
\hskip0.35cm
  \includegraphics[width=13cm, height = 6cm]{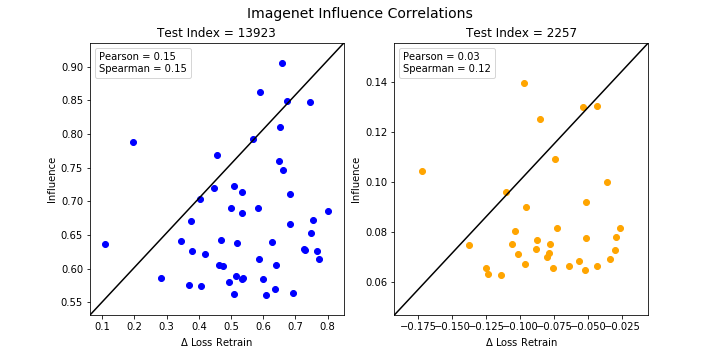}
  \caption{\label{imagenet}   ImageNet influence estimation results for the selected test points 13,923 (left) and 2,257 (right). X-axis is change in test loss after removal of a training point and retraining as described in the text. Y-axis is the change in test loss estimated with influence function. Pearson and Spearman correlations are shown in the caption. Correlations are low, showing the weakness of this influence estimation.%\myred{we need a better caption, explain it. also make the figure a bit smaller} 
  }
\end{figure*}

\begin{figure*}[ht]
%\hskip-0.8cm
  \includegraphics[width=14cm]{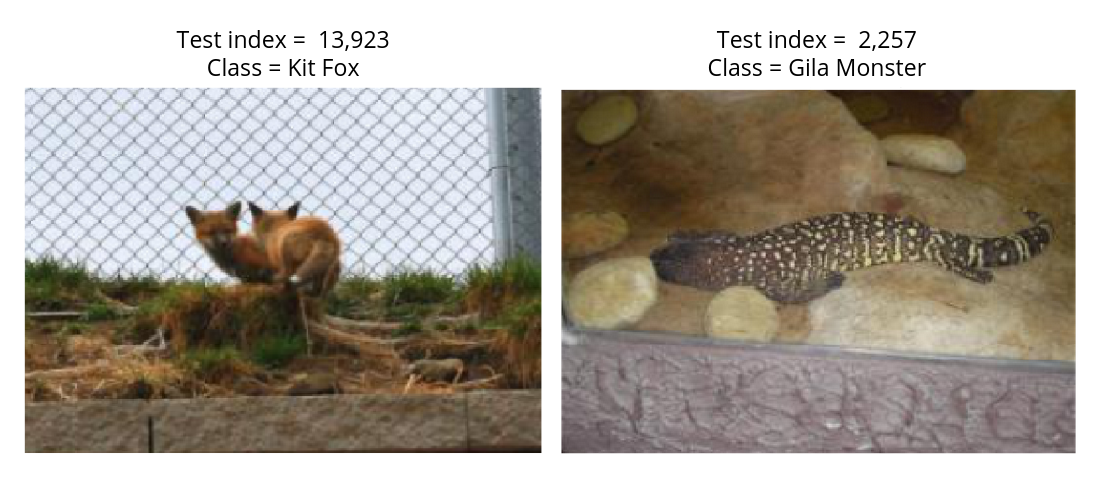}
  \caption{\label{imagenet-test} 
  Selected test points for influence estimation.
  %\myred{make the fig a bit smaller. also two panels don't have same sizes}  
  }
\end{figure*}

\begin{figure*}[ht]
\hskip-0.6cm
  \includegraphics[width=16cm]{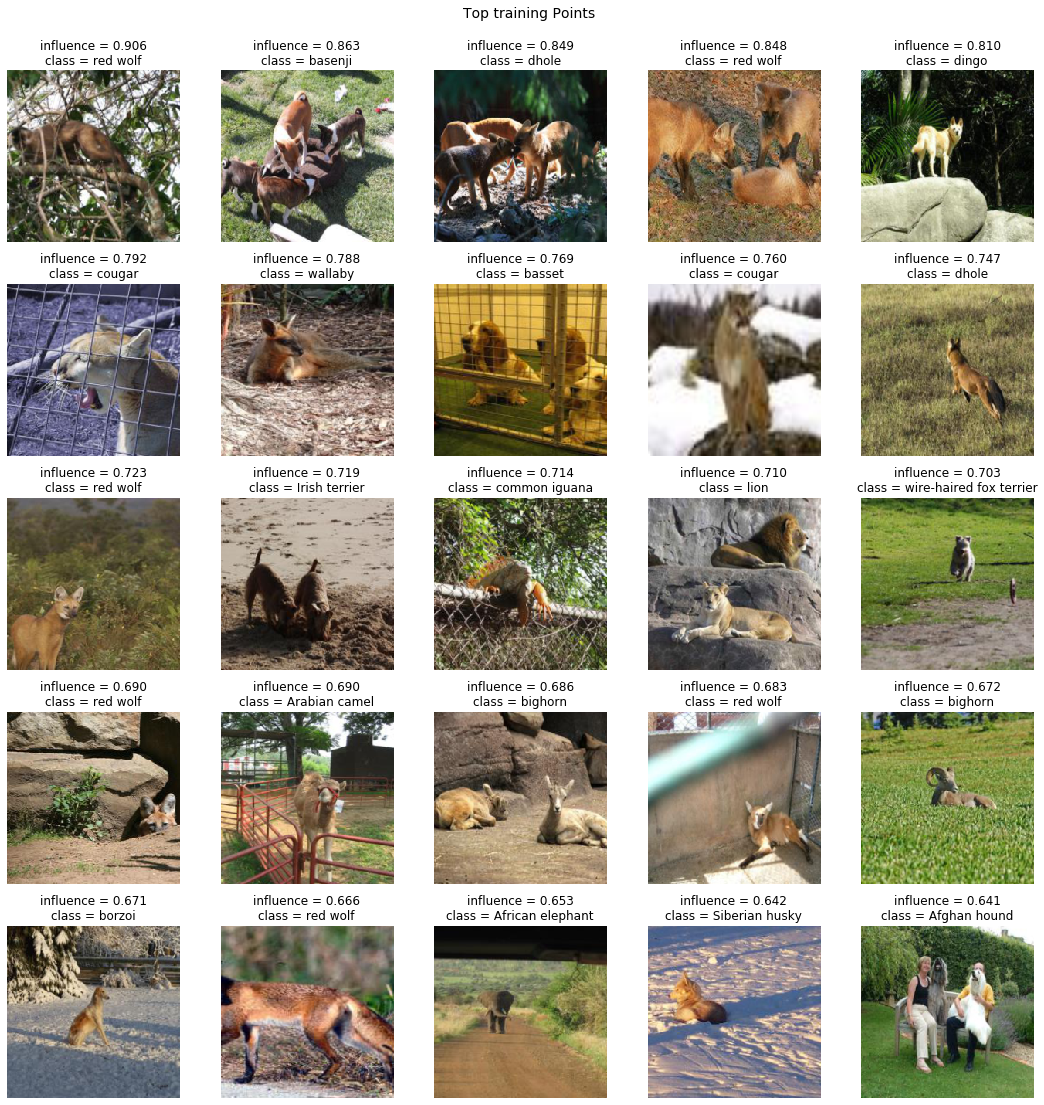}
  \caption{\label{13293-train} Top 25 ImageNet training points by influence for test point 13,293, kit fox. 
  Many of the identified classes are furred mammals, e.g. red wolf, basenji, and dingo, which have visual similarity to the test point. Other examples are questionable, e.g. the common iguana, and African elephant. Although there is qualitative similarity is some cases, the results are still overall weak quantitatively.
  %\myred{this message is contradicting with our other arguments that influence estimates are not reliable. Without a quantifiable statistics, it is hard to argue qualitatively from these figures that influence estimates are good. we can highlight both good and bad examples. Same comments on the next figure and also in the text.  }
  }
\end{figure*}

\begin{figure*}[ht]
\hskip-0.6cm
  \includegraphics[width=16cm]{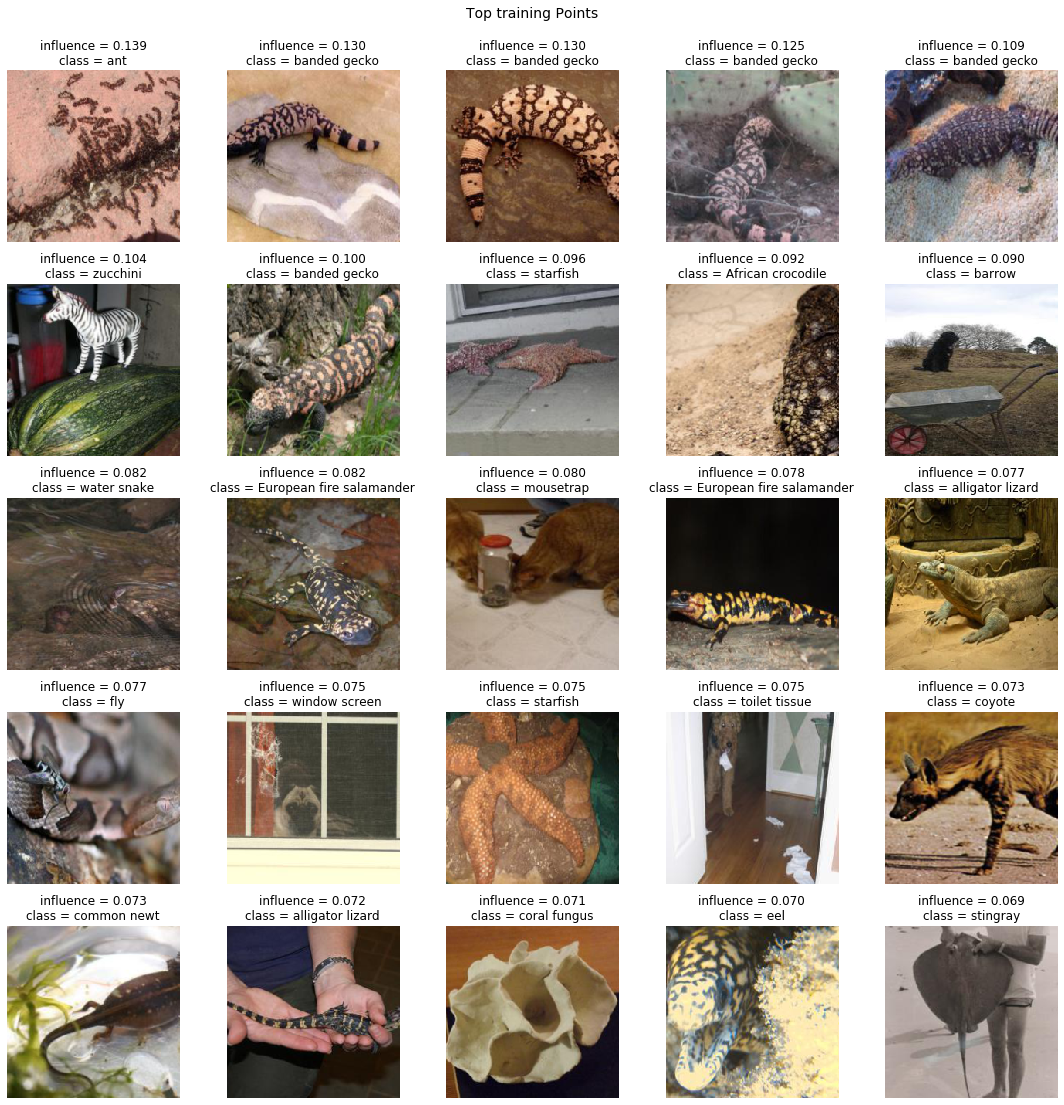}
  \caption{\label{2257-train} Top 25 ImageNet training points by influence for test point 2,257, gila monster.   Many of the identified classes are spotted lizards, e.g. banded gecko ad European fire salamander, which have visual similarity to the test point. Other examples are questionable, e.g. the stingray, coral fungus, and barrow. Although there is qualitative similarity is some cases, the results are still overall weak quantitatively.
  } 
\end{figure*}

We plot the obtained correlations in Figure \ref{imagenet}.

For computing the weight gradient norm, we take the mean norm in batches of size 128 over the entire dataset for both our model and a standard PyTorch pretrained model as a baseline, both of which are ResNet-50 models with around 25.5M parameters.

\newpage
\subsection{computing inverse-Hessian vector product}
\label{sec:hessian-vector}
In large over-parameterized deep networks, computing and inverting the exact Hessian $H_{\btheta^{*}}$ is expensive. In such cases, the Hessian-vector product rule \citep{Pearlmutter} is used along with conjugate-gradient \citep{conjugate_gradient} or stochastic estimation \citep{Agarwal2016SecondOS} to compute the approximate inverse-Hessian Vector product. More specifically, to compute $t = H_{\btheta^{*}}^{-1}v$, we solve the following optimization problem using conjugate-gradient: 
    $
    t^{*} = \arg\min_{t} \{ \frac{1}{2} t^{T}H_{\btheta^{*}}t - v^{T}t \}$,
where $v = \nabla_{\btheta} \ell(h_{\btheta^{*}}(z_{t}))$. This optimization, however, requires the Hessian $H_{\btheta^{*}}$ to be a positive definite matrix, which is not true in case of deep networks due to the presence of negative eigenvalues. In practice, the Hessian can be regularized by adding a damping factor of $\lambda$ to its eigenvalues (i.e. $H_{\btheta^{*}} + \lambda I$) to make it positive definite.

In deep models, with a large number of parameters and large training set, conjugate-gradient is often expensive as it requires computing the Hessian-vector product \citep{Pearlmutter} for every data sample in the training set. In those cases, stochastic estimation techniques \citep{Agarwal2016SecondOS} have been used which are fast as they do not require going through all the training samples. In stochastic estimation, the inverse Hessian is computed using a recursive reformulation of the Taylor expansion: $H_{j}^{-1} = I + (I-H)H_{j-1}^{-1}$ where $j$ is the recursion depth hyperparameter . A training example $z_{i}$ is uniformly sampled and $\nabla^{2} \ell(h_{\btheta^{*}}(z_{i}))$ is used as an estimator for computing $H$. This technique also requires tuning a scaling hyperparameter $\gamma$ and a damping hyperparameter $\beta$ 
\footnote{It is assumed that $\forall i, I - \nabla^{2} \ell(h_{\btheta^{*}}(z_{i})) \succcurlyeq 0$; \citep{influence1} notes that if this is not true, the loss can be scaled down without affecting the parameters. The scaling factor is a hyperparameter which helps the convergence of the Taylor series. The damping coefficient is added to the diagonal of the Hessian matrix to make it invertible.}. In our experiments with large deep models, we use the stochastic estimation method to compute the inverse-Hessian Vector product.
\subsection{Effect of Initialisation and Optimizers on Influence Estimates}
To understand the effect of network initialisation on the quality of influence estimates, we compute the influence scores across different random initialisations. The influence estimates are computed for the small CNN architecture \citep{influence1} and LeNet \citep{lenet}, both trained on the MNIST dataset. Both the architectures are trained with a constant weight-decay factor of 0.001. \begin{wrapfigure}{r}{0.5\textwidth}
\vspace{-1.8em}
  \begin{center}
    \includegraphics[width=0.5\textwidth]{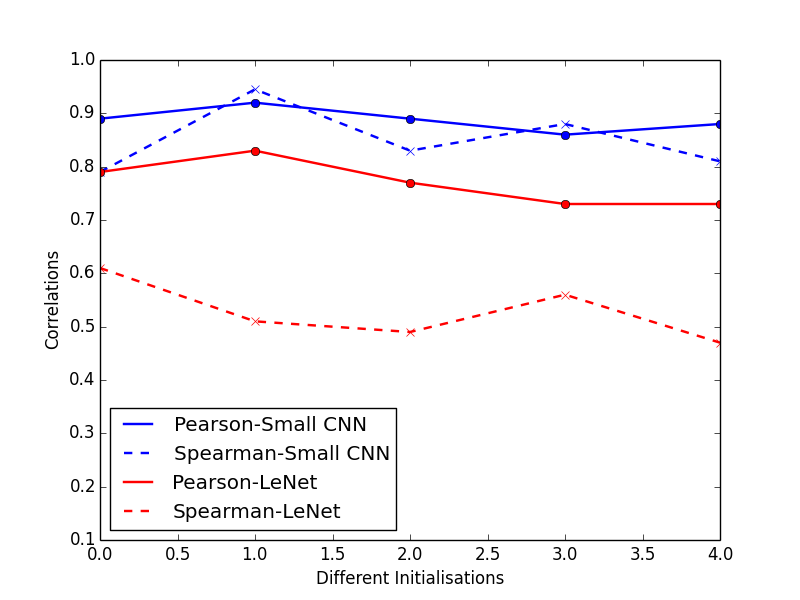}
  \end{center}
  \vspace{-1.5em}
  \caption{\label{initialisation_exp}Correlations with different network initialisation}
\end{wrapfigure}
In Fig \ref{initialisation_exp}, we observe that across different network initialisations, although both the Pearson and Spearman correlations between the influence estimates and the ground-truth are inconsistent, the variance amongst them is particularly low. Note that for both the network architectures, we compute the influence estimates for the test-point with the highest loss at the optimal model parameters. The correlation between the influence estimates and leave-out re-trainings are computed with the top 40 influential training examples. Additionally to understand the impact of the selection of optimizer on the influence estimates, we train the LeNet architecture on MNIST with different optimizers namely Adam \citep{kingma2014adam}, Gradient Descent \citep{Bottou10large-scalemachine}, Nesterov and RMSProp \citep{DBLP:journals/corr/Ruder16}.  We notice that the Pearson correlation $(0.72 \pm 0.04$) has a marginally lower variance when compared to the Spearman rank-order correlation ($0.56 \pm 0.11)$.

\subsection{Effect of training sample selection for ground-truth influence}
In this section, we understand the effect of selecting different number of training samples on the correlation estimates. We investigate this with a case study for a CNN architecture trained on small MNIST. Keeping a test-point with a high loss fixed, we sample different sets of training examples with the highest and the lowest influence scores over different network initialisations.  \begin{wrapfigure}{r}{0.55\textwidth}
\vspace{-1em}
\hskip-10em
  \begin{center}
    \includegraphics[width=0.55\textwidth]{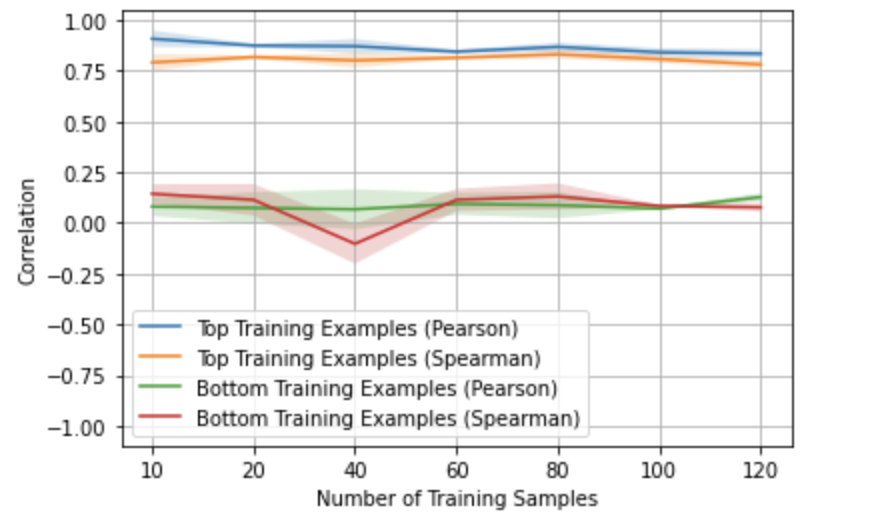}
  \end{center}
  \vspace{-1em}
  \caption{\label{training_sample}Correlations with different training samples}
  \vspace{-2em}
\end{wrapfigure}
Note that in this setting as shown in the main paper, the quality of influence estimates are relatively good. We observe that when the influence estimates are evaluated with the top influential points, both the Pearson and Spearman correlations are relevant. This is true across different number of training samples. However when the evaluation carried out with respect to the lowest influential training samples, the correlation estimates are of poor quality. These results highlight the importance of the selection of the type of training samples with respect to which the correlation estimates are computed. 
\subsection{Faithfulness and Plausibility of Influence functions}
The authors \citep{Jacovi_2020} primarily tackle the importance and trade-offs between plausibility (i.e. if the interpretations are convincing to humans) and faithfulness (i.e. how accurate an interpretation is to the “true reasoning process of the model”) of existing interpretation methods. To the best of our knowledge such an analysis has not been done for influence functions. We observe that explanations from influence functions for deep networks are sometimes plausible and sometimes not. For instance, in Appendix Fig. \ref{cifar_1}, we observe that the selection of test-point with (class = bird) leads to training examples with (class = deer) amongst the top influential points. On the other hand, in Appendix Fig. \ref{cifar_2}, we observe many plausible explanations. Influence functions that work are faithful because they answer the following question:``what would this model have done if certain data were excluded?”. This class of questions, while not exhaustive, have special relevance because they are counterfactuals, which hold both intuitive appeal and for their special status in causal reasoning. However, we must be cautious because they may not be faithful when they incur approximation errors, as highlighted in our paper.
\subsection{CIFAR-100 Influential Examples}
\begin{figure*}[ht]
  \includegraphics[width=14cm]{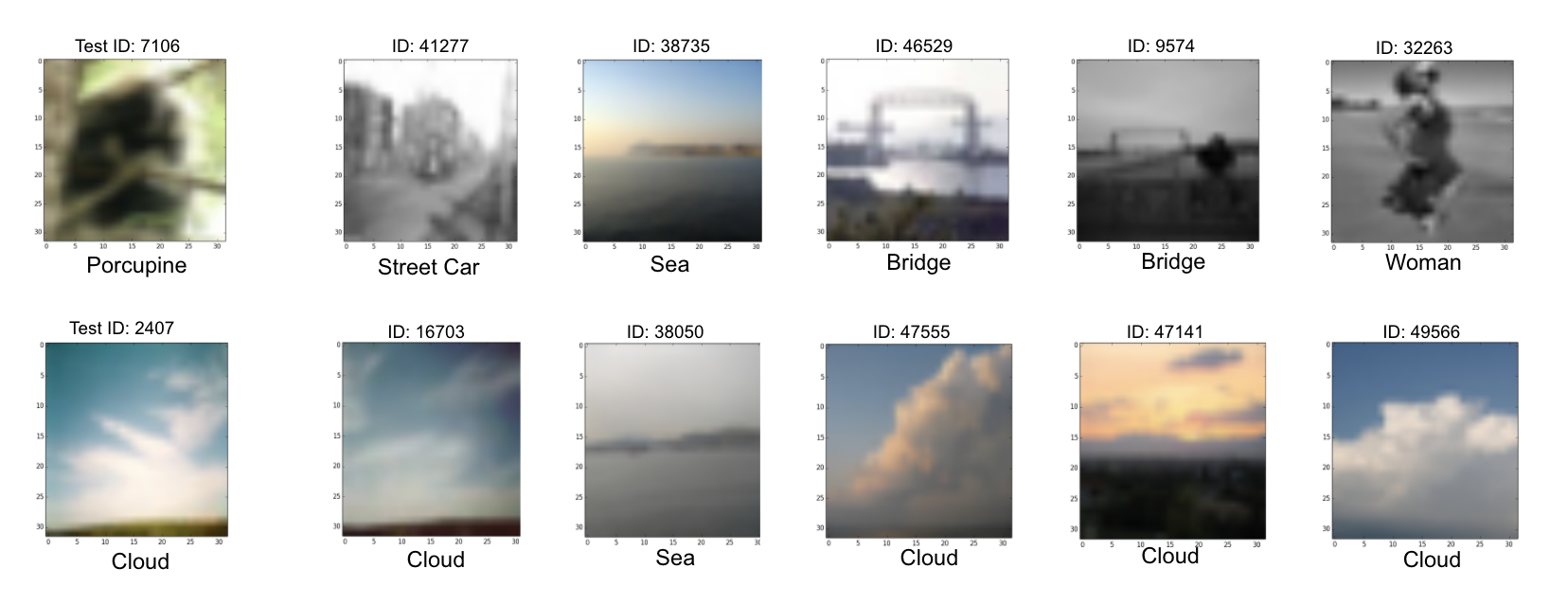}
  \caption{\label{cifar_100_examples}Top 5 influential points for the test points: 7106 and 2407 (CIFAR-100). The model is a ResNet-18 trained with a weight-decay regularization. For the test-point with index 7106, the influential training samples are semantically dissimilar from the test-point. However for the test-point with index 2407, 4 out of the top 5 samples share semantic similarity with the test-point. }
\end{figure*}
\subsection{Preliminary Results on Group Influence}
\begin{wrapfigure}{r}{0.55\textwidth}
\vspace{-1em}
\hskip-10em
  \begin{center}
    \includegraphics[width=0.55\textwidth]{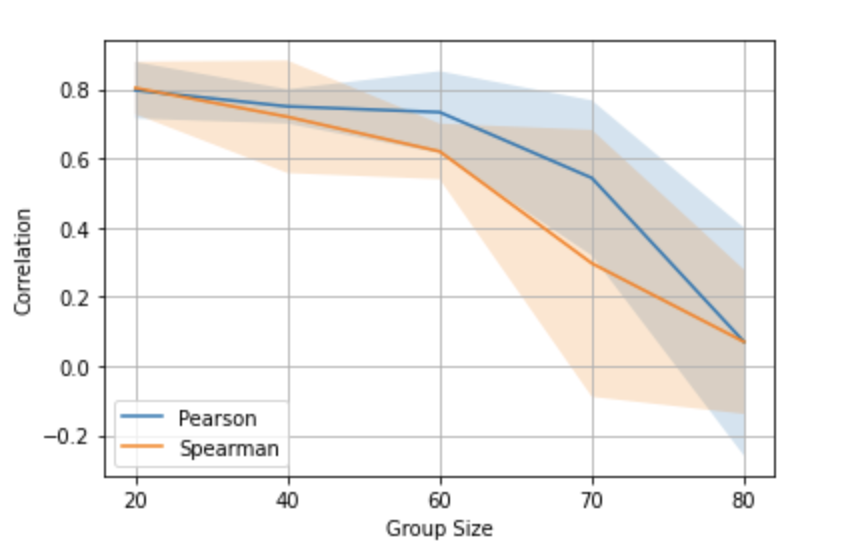}
  \end{center}
  \vspace{-1em}
  \caption{\label{groups}Group Influence on Iris}
  \vspace{-1em}
\end{wrapfigure}
Understanding model changes when a group of training samples are up-weighted is indeed an important research problem. Influence functions \citep{cook_influence, influence1} in general are accurate when the model perturbation is small. However when a group of samples are up-weighted, the model perturbation is large, which violates the small perturbation assumption of influence functions. Previously it has been shown \citep{influence2, Basu2019SecondOrderGI} that group influence functions are fairly accurate for linear and convex models, even when the model perturbation is substantial. In this section, we present some preliminary results on the behaviour of group influence functions for non-convex models. Our main observation is that group influence functions are fairly accurate for small networks. Nonetheless for large and complex networks, the influence estimates are of poor quality. For e.g. in Fig. \ref{groups}, we observe that the correlation estimates for small group sizes are accurate, whereas for larger group sizes, the estimates are of poor quality. For a ResNet-18 model trained on MNIST (with a weight-decay regularization factor of 0.001), we observe the correlation estimates across different group sizes to vary from 0.01 to 0.21. Similarly for a ResNet-18 trained on CIFAR-100, we observe the group influence correlation estimates to range from 0.01 to 0.18. We leave the complete investigation of group influence in deep learning as a direction for future work.
\subsection{Additional experiments with Multiple Test-Points}
In our experimental setup, we evaluate the correlation estimates with respect to one test-point at a time. Although the evaluation of the correlation estimates with multiple test-points is more robust, it comes at the expense of high computational cost. To illustrate the quality of influence estimates with multiple test-points, we compute the influence estimates for small MNIST with 8 different test-points.  We sample two test-points each from : (a) $100^{th}$ percentile of the test-loss; (b) $75^{th}$ percentile of the test-loss; (c) $50^{th}$ percentile of the test-loss; (d) $25^{th}$ percentile of the test-loss. The Pearson and Spearman correlations are 0.91 and 0.78 respectively. In a similar setting, for a complex architecture such as ResNet-18 trained on CIFAR-100, the Pearson and Spearman correlations are 0.15 and 0.11 respectively.
\begin{figure}
\begin{center}
\includegraphics[width=26em]{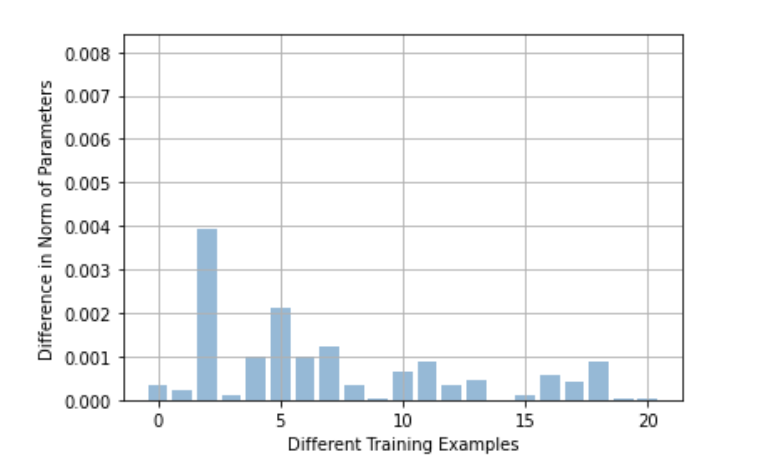}
\caption{Norm of difference in parameters obtained by training from scratch vs. re-training from optimal parameters }
\end{center}
\end{figure}
\begin{figure}
\begin{center}
\hspace{-2.5em}
\includegraphics[width=26em]{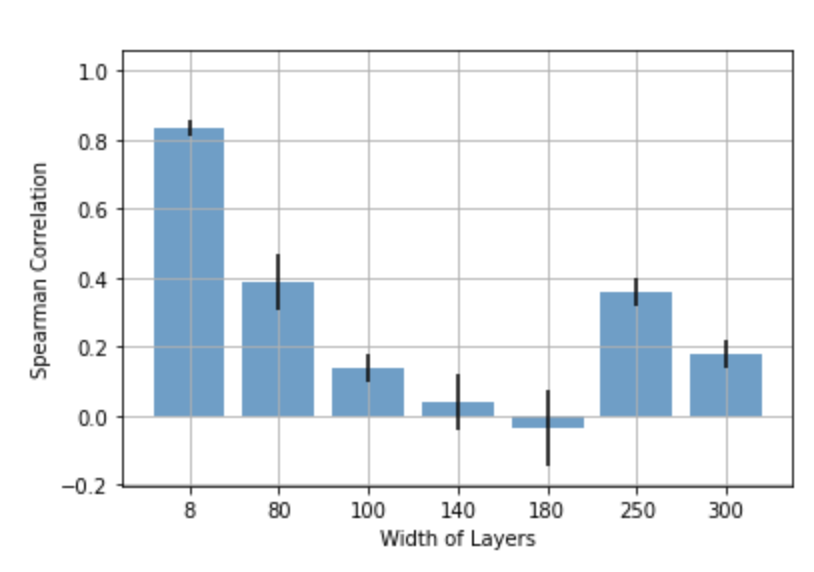}
\caption{Width vs. Spearman Correlation for a one-layered network }
\end{center}
\end{figure}
\subsection{Impact of Activation Functions}
In our experiments we observe that even with non-smooth activation functions such as ReLU, we obtain high quality influence estimates for certain networks. Understanding influence estimates with ReLU has an additional challenge since there measure zero subsets where the function is non-differentiable.  Recently \citep{serra2018bounding} has provided improved bounds on the number of linear regions for shallow ReLU networks. Understanding the impact of the number of linear regions in ReLU networks on the influence estimates is an interesting research direction, however we defer it for future work.

%% Please note that we have introduced automatic line number generation
%% into the style file for \LaTeXe. This is to help reviewers
%% refer to specific lines of the paper when they make their comments. Please do
%% NOT refer to these line numbers in your paper as they will be removed from the
%% style file for the final version of accepted papers.

%\subsubsection*{Author Contributions}
%If you'd like to, you may include  a section for author contributions as is done in many journals. This is optional and at the discretion of the authors.

%\subsubsection*{Acknowledgments}
%Use unnumbered third level headings for the acknowledgments. All
%acknowledgments, including those to funding agencies, go at the end of the paper.

\end{document}